\pdfoutput=1
\documentclass[runningheads]{llncs}
\usepackage[T1]{fontenc}
\usepackage[utf8]{inputenc}
\usepackage{graphicx}
\usepackage{hyperref}
\usepackage[svgnames]{xcolor}

\usepackage{booktabs}
\usepackage{makecell}
\usepackage[export]{adjustbox}
\usepackage{microtype}
\usepackage{multirow}

\usepackage{fancyhdr}
\fancypagestyle{officialbibref}{\fancyhf{}\fancyheadoffset[L,R]{50pt}\fancyhead[C]{\kern-10em This paper was published at \textbf{TSD 2025}, please cite the published version {\small\url{https://doi.org/10.1007/978-3-032-02551-7_7}}.\kern-10em}\fancyfoot[C]{}\addtolength{\topmargin}{-82pt}\addtolength{\headsep}{82pt}}
\begin{document}
\title{Refining Czech GEC: Insights from a~Multi-Experiment Approach}

\titlerunning{Refining Czech GEC}
\author{Petr Pechman\inst{1}\orcidID{0009-0009-7512-0115} \and
Milan Straka\inst{2}\orcidID{0000-0003-3295-5576}
\and
Jana~Straková\inst{2}\orcidID{0000-0003-0075-2408}
\and
Jakub Náplava\inst{1}\orcidID{0000-0003-2259-1377}}

\authorrunning{P. Pechman et al.}

\institute{Seznam.cz, Prague, Czech Republic\\
\email{\{petr.pechman,jakub.naplava\}@firma.seznam.cz}
\and
Charles University, Faculty of Mathematics and Physics, Institute of Formal and Applied Linguistics, Prague, Czech Republic \\
\email{\{straka,strakova\}@ufal.mff.cuni.cz}
}

\maketitle              %
\thispagestyle{officialbibref}
\begin{abstract}
We present a grammar error correction (GEC) system that achieves state of the art for the Czech language. Our system is based on a neural network translation approach with the Transformer architecture, and its key feature is its real-time synthetic generation pipeline, which dynamically augments sentences with artificial errors by introducing both language-agnostic and Czech-specific errors. We conduct a comprehensive series of experiments, investigating the Czech GEC corpora as bases for synthetic error introduction, several error generation strategies, domain balancing, tokenization granularity, model size, and data scaling during fine-tuning.  Additionally, we evaluate the performance of large language models (LLMs) on Czech GEC in both end-user and expert fine-tuning scenarios. Our best-performing model is superior both in performance and computational efficiency. The source code and the trained model links are available on \url{https://github.com/ufal/tsd2025-gec}.

\end{abstract}

\section{Introduction}

Grammar error correction (GEC) is a task that aims to correct errors
in natural texts. The errors can range from simple typos and spelling errors to more
complex issues such as incorrect verb tenses, subject-verb agreement errors, punctuation errors, and syntactic inconsistencies.

In this work, we publish a state-of-the-art Transformer-based neural machine translation Czech grammar error correction system trained on a mixture of synthetic and manually annotated data. The key feature of our system is a real-time synthetic data augmentation pipeline, capable of generating both language-agnostic and Czech-specific errors as synthetic data for learning.

We conduct an extensive series of experiments, and we present our insights drawn from the results. Our experiments focus on corpora selection for pre-training and fine-tuning, domain balancing, tokenization granularity, model size, and data scaling during fine-tuning data.

Furthermore, we evaluate the large language models (LLMs), as well as available Czech grammar error correction tools on the Czech GEC task in the following scenarios: We present a quantitative evaluation of an end-user use-case with ready-to-use GEC tools, including off-the-shelf LLMs. Additionally, we employ LLMs fine-tuned to the manually annotated GEC data, and even LLMs that underwent continued pre-training for 
 Czech in general and then were fine-tuned to grammar error correction.

To conclude, we release two models, each fine-tuned with one of the two publicly available manually annotated Czech GEC datasets: a model fine-tuned with the AKCES-GEC corpus \cite{naplava-straka-2019-low}, and a model fine-tuned with the Grammar Error Correction  Corpus, the GECCC \cite{naplava-etal-2022-geccc}. The latter, our best-performing model, achieves state-of-the-art results in Czech using a fraction of computational resources required by LLMs.

\section{Related Work}
\label{sec:related-work}

\subsubsection*{Czech Grammar Error Correction Tools} \textit{Korektor} \cite{richter-etal-2012-korektor} is a HMM-based Czech text correction tool, which corrects typos and diacritics, but is limited to word corrections only, i.e., it cannot add or remove a word. \textit{Opravidlo} \cite{hlavackova-etal-2022-opravidlo} is a rule-based system using five different rule modules, each targeting a specific range of errors. It also uses an extensive dictionary for correction of typos. Also, various word processors, such as \textit{MS Word} or \textit{Google Docs}, offer spelling and grammar checking. Last, but not least, generative AI tools based on large language models (LLMs), such as ChatGPT,\footnote{\url{https://chatgpt.com/}} or the open-sourced DeepSeek \cite{deepseekai2025short} can also be used for text correction.

\subsubsection*{Related Methods} Pre-training a Transformer-based GEC system \cite{grundkiewicz-etal-2019-neural} using synthetic data was previously shown by Náplava~\&~Straka~(2019)~\cite{naplava-straka-2019-low}. They also further fine-tuned their system on a mixture of Czech synthetic and manually annotated GEC data (AKCES-GEC) to avoid overfitting and improve performance. Náplava~et~al.~(2022) fine-tune the same architecture on a mixture of synthetic data and GECCC, a large manually annotated GEC corpus \cite{naplava-etal-2022-geccc}.

\section{Data}

For Czech, two manually annotated GEC datasets are available: AKCES-GEC \cite{naplava-straka-2019-low}, and the Grammar Error Correction Corpus for Czech (GECCC; \cite{naplava-etal-2022-geccc}). The latter is further divided into four distinct domains: Natives Formal (NF), Natives Web Informal (NWI), Romani (R), and Second Learners (SL).

For synthetic data generation, we follow an unsupervised approach \cite{naplava-straka-2019-low,grundkiewicz-etal-2019-neural} for introducing artificial errors into the following monolingual Czech corpora: the Czech subset of the CoNLL 2017 Shared Tasks texts from Common Crawl \cite{corpus-common-crawl-2017}, the SYN v4 corpus \cite{hnatkovasyn}, the News 2019 corpus \cite{barrault-etal-2019-findings}, and the Wikipedia corpus presented within DaMuEL \cite{kubesa-and-straka-2023-wikipedia}.%

\section{Methodology}

\begin{table}[t]
    \centering
    \caption{Probability distribution of token-level noising operations.}
    \label{tab:token-level-ops-and-typical-errors}
    \setlength{\tabcolsep}{4.8pt}
    \renewcommand\cellset{\renewcommand\arraystretch{0.75}}

    \begin{tabular}{l|cccccc|c}
    \toprule
     & \multicolumn{6}{ c |}{Token-level Operations} & Typical Errors \\ 
    \midrule 
    \multicolumn{1}{l|}{Setting} & \makecell[c]{sub \\ Aspell} & \makecell[c]{sub \\ MorphoDiTa} & ins & del & swap & recase & \\
    \midrule
    \textit{Aspell} & 0.7 & 0.0  & 0.1 & 0.05 & 0.1 & 0.05 & No \\
    \textit{MorphoDiTa} & 0.5 & 0.2  & 0.1 & 0.05 & 0.1 & 0.05 & No \\
    \textit{Typical Errors} & 0.7 & 0.0  & 0.1 & 0.05 & 0.1 & 0.05 & Yes \\
    \textit{MATE} & 0.5 & 0.2  & 0.1 & 0.05 & 0.1 & 0.05 & Yes \\
    \bottomrule
    \end{tabular}
\end{table}

Our system follows the neural translation approach to GEC using a Transformer model, first pre-trained in an unsupervised fashion on a mix of clean and synthetic data, and then fine-tuned on either the AKCES-GEC or GECCC corpus, optionally augmented with additional synthetic errors.

\subsubsection*{Data Generation}

For lack of large amount of manually annotated GEC data sufficient for training the Transformer, we synthetically augment raw, monolingual training data by injecting errors. Our pipeline generates the perturbed data in real time during training. First, artificial, language-agnostic errors are introduced into a corpus; then, we add typical Czech errors.

First, for each sentence, we sample a probability from a normal distribution (in which mean and standard deviation are hyperparameters of the method), multiplying this sampled constant with a number of words in the sentence to determine the number of words to be modified. Then, on the selected words, one of the character-level operations \textit{substitution}, \textit{insertion}, \textit{deletion}, \textit{swap} or \textit{diacritics} is applied using a uniform probability distribution over these operations. The word for the \textit{substitution} operation is suggested by either Aspell or MorphoDiTa \cite{strakova-etal-2014-morphodita}. The same process is applied for token-level operations, with the corresponding probability distribution shown in Table~\ref{tab:token-level-ops-and-typical-errors}.

\looseness-1
In the second step, we optionally add typical Czech errors (as indicated in Table~\ref{tab:token-level-ops-and-typical-errors}),
regardless of whether the word has already been perturbed in the previous step. For typical Czech errors, we have identified common grammatical errors from the manually annotated corpora, e.g., \textit{mě/mně}, incorrect capitalization, punctuation errors, \textit{i/y}, \textit{s-/z-}, \textit{ú/ů}, \textit{bysme$^*$/bychom}, \textit{dvouma$^*$/dvěma}, etc.

We call a combination of the above-mentioned operations \textbf{MATE}, standing for token substitutions both by \textbf{M}orphoDiTa and \textbf{A}spell, and \textbf{T}ypical \textbf{E}rrors.

\subsubsection*{Evaluation}

Quite a few evaluation measures have been proposed for grammar error correction. We use the M\textsuperscript{2} Scorer \cite{dahlmeier-ng-2012-m2}, which has been shown to correlate the most with the human judgement on the GECCC dataset~\cite{naplava-etal-2022-geccc}.

Unless stated otherwise, we always report pre-training results after a fixed number of epochs, and fine-tuning results using the checkpoint from the best-performing epoch based on the development data.

\section{Results}
\label{sec:results}

\subsubsection*{Corpora Used for Pre-Training and Fine-Tuning} 
\begin{figure}[t]
    \begin{minipage}[t]{.49\hsize}
    \centering
    \begin{tabular}{ lcc }
    \toprule
    \multicolumn{1}{c}{Corpus} & AKCES & GECCC \\
    \midrule
    Common Crawl & 65.92 & 55.27 \\
    News 2019 & 71.88 & 63.54 \\
    SYN-v4 & \textbf{72.79} & \textbf{64.00} \\
    Wikipedia & 70.63 & 60.23 \\
    \bottomrule
    \end{tabular}
    \end{minipage}
    \begin{minipage}[t]{.5\hsize}
    \centering
    \includegraphics[width=\hsize,valign=M,raise=-.05cm]{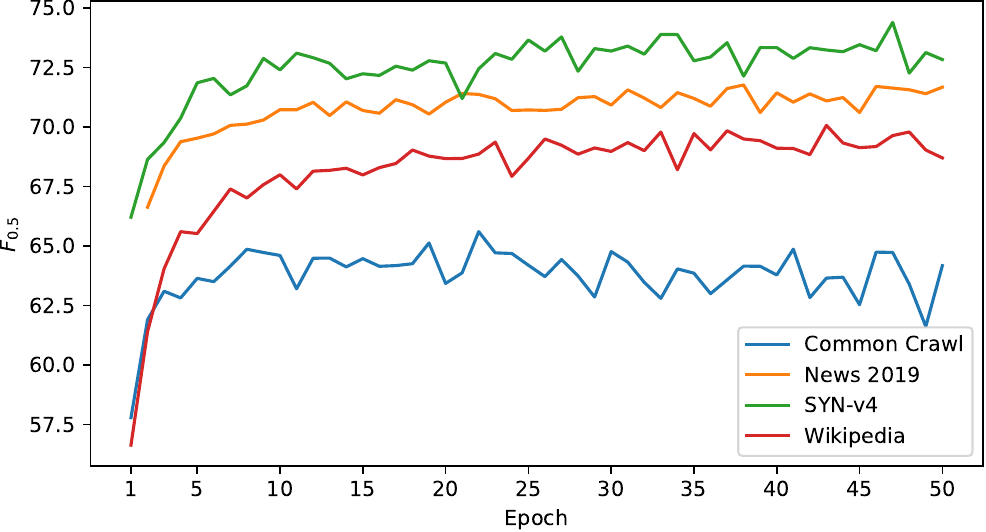}
    \end{minipage}
    
    \caption{Comparison of the contribution of synthetic error source corpora after pre-training (50\textsuperscript{th} epoch). Left: AKCES and GECCC test sets (F\textsubscript{0.5}-score). Right: training progress evaluated on the GECCC development set.}
    \label{fig:corpora-comparison-pretraining}
\end{figure}

Fig.~\ref{fig:corpora-comparison-pretraining} compares a selection of Czech monolingual corpora used as bases for synthetic error introduction during pre-training. Table~\ref{tab:corpora-comparison-finetuning} shows models pretrained with the distinct synthetic-error source corpora, followed by fine-tuning on manually annotated target data.
We attribute the lowest performance of Common Crawl to its relatively high noisiness, while the performance of the other corpora corresponds to their size (9.7M, 14.6M, and 28.6M sentences for Wikipedia, News 2019, and SYN-v4, respectively). %

\begin{table}[t]
    \centering
    \caption{Results of fine-tuning on manually annotated data after pretraining on different corpora with synthetic errors (F\textsubscript{0.5}-score).}
    \label{tab:corpora-comparison-finetuning}
    
    \begin{tabular}{ l|c|ccccc }
    \toprule
    \multicolumn{1}{c|}{Corpus} & AKCES & GECCC & NF & NWI & R & SL \\
    \midrule
    Common Crawl & 79.73 & 71.93 & 69.83 & 75.92 & 72.28 & 69.35 \\
    News 2019 & 79.76 & 72.52 & 70.52 & 75.53 & \textbf{72.80} & 70.66 \\
    SYN-v4 & \textbf{81.32} & \textbf{72.97} & \textbf{72.70} & \textbf{76.29} & 72.78 & \textbf{70.83} \\
    Wikipedia & 79.25 & 70.94 & 68.38 & 73.91 & 71.18 & 69.15 \\
    \bottomrule
    \end{tabular}
\end{table}

\subsubsection*{Data Generation} 
\begin{figure}[t]
    \begin{minipage}[t]{.49\hsize}
    \centering
    \begin{tabular}{lcc}
    \toprule
    \multicolumn{1}{c}{Setting} & AKCES & GECCC \\
    \midrule
        Aspell & 67.22 & 55.41 \\
        MorphoDiTa & 64.19 & 53.24 \\
        Typical Errors & 73.58 & \textbf{65.55} \\
        MATE & \textbf{74.23} & \textbf{65.55} \\        
        \bottomrule
    \end{tabular}
    \end{minipage}
    \begin{minipage}[t]{.5\hsize}
    \centering
    \includegraphics[width=\hsize,valign=M,raise=-.05cm]{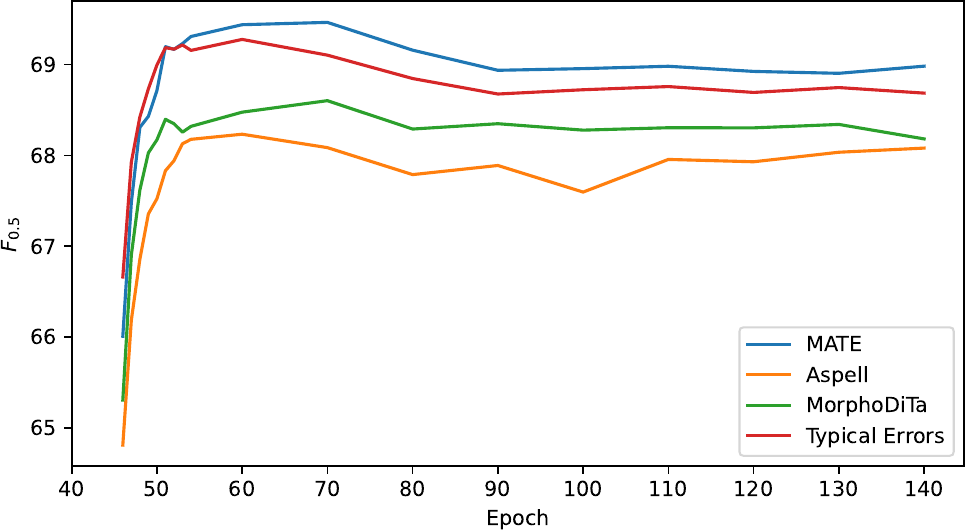}
    \end{minipage}

    \caption{The effect of introducing errors from various sources during pre-training (stopped after the 45\textsuperscript{th} epoch). Left: results immediately after pre-training on the AKCES and GECCC test sets (F\textsubscript{0.5}-score). Right: subsequent fine-tuning progress evaluated on the GECCC development set.}
    \label{fig:finetuning-geccc}
\end{figure}

Fig.~\ref{fig:finetuning-geccc} shows the effect of introducing errors from various sources during pre-training, evaluated after 45\textsuperscript{th} epoch, and learning progress during the subsequent fine-tuning phase, clearly proving that a combination of methods (MATE) outperforms all the individual ones.

So far, we have presented results where the model was pre-trained on corpora with synthetic errors and subsequently fine-tuned using only manually annotated data. 
Following Náplava et al. (2019) \cite{naplava-straka-2019-low}, we now evaluate the effect of incorporating synthetic errors into the fine-tuning stage.

\begin{figure}[t]
    \begin{minipage}[t]{.49\hsize}
    \centering
    \begin{tabular}{ccc}
    \toprule
    \makecell[c]{Synthetic\\Data Ratio} & AKCES & GECCC \\
    \midrule
    0:1 & \textbf{81.19} & \textbf{72.25} \\
    2:1 & 80.98 & 72.08 \\
    5:1 & 80.24 & 71.86 \\
    \bottomrule
    \end{tabular}    
    \end{minipage}
    \begin{minipage}[t]{.5\hsize}
    \centering
    \includegraphics[width=\hsize,valign=M,raise=-.05cm]{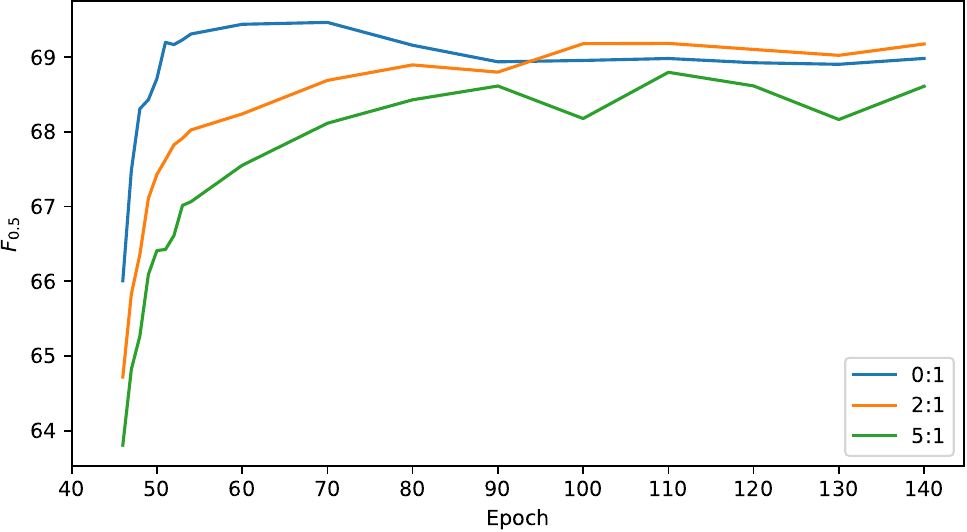}
    \end{minipage}

    \caption{Synthetic data vs. manually annotated data ratios using the combined \textit{MATE} method during fine-tuning. Left: final results after fine-tuning on the AKCES and GECCC test set (F\textsubscript{0.5}-score). Right: fine-tuning progress evaluated on the GECCC dev set.}
    \label{fig:finetuning-mate-ratios}
\end{figure}

Fig.~\ref{fig:finetuning-mate-ratios} displays the results when mixing synthetic and manually annotated data during the fine-tuning phase. Surprisingly, our optimal setting avoids using synthetic errors completely during fine-tuning, which is contrary to the findings of Náplava~et~al.~(2019) \cite{naplava-straka-2019-low}, who combined synthetic and manually annotated data in a 2:1 ratio during fine-tuning.

\subsubsection*{Domain Balancing and Domain-Specific Models} 

To maximize performance on a selected domain of the GECCC, we ask whether it is more beneficial to fine-tune exclusively on the target domain rather than on the entire corpus. Such comparison is presented in Table~\ref{tab:relative-improvements-on-domains}. Fine-tuning specifically for the target domain brings relative improvement against the models trained generally on the entire GECCC, despite the general models having seen more data.

\begin{figure}[t!]
    \centering
    \includegraphics[width=.7\hsize]{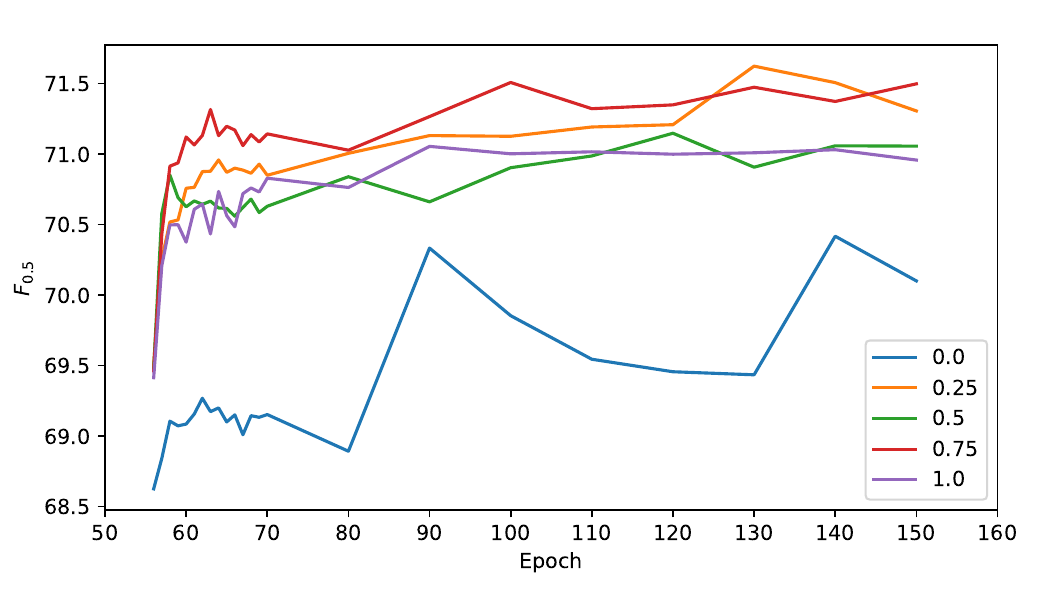}
    \caption{Fine-tuning progress using oversampling factors evaluated on GECCC dev set.}
    \label{fig:finetuning-geccc-oversampling}
\end{figure}

\looseness1
Now, we focus on maximizing the overall score across all domains. As each of the differently-sized domains contributes unevenly to the final score, we adopt a controlled (over)sampling strategy: each domain is sampled in training according to $\mathit{size}^\mathit{factor}$, where $\mathit{size}$ is the domain size and $\mathit{factor}$ is a hyperparameter, see Table~\ref{tab:finetuning-geccc-oversampling} and Fig.~\ref{fig:finetuning-geccc-oversampling} for results. The conclusion from the presented results is that carefully balancing domain (over)sampling is beneficial for the overall score.

\begin{table}[t]
    \centering
    \caption{The relative improvement of a domain-specific fine-tuning evaluated with respect to the performance of the model fine-tuned on the entire GECCC.}
    \label{tab:relative-improvements-on-domains}
    \begin{tabular}{l|l|c|c}
    \toprule
    \multicolumn{1}{c|}{Evaluation Data}  & \multicolumn{1}{c|}{Fine-tuning Data}  & F\textsubscript{0.5}-score & \shortstack{Relative \\ Improvement} \\
    \midrule
    Natives Formal & GECCC & 70.46 & \\
    Natives Formal & Natives Formal & 71.87 & +1.41 \\
    \midrule
    Natives Web Informal & GECCC & 75.91 & \\
    Natives Web Informal & Natives Web Informal & 75.83 & -0.08 \\
    \midrule
    Romani & GECCC & 73.04 & \\
    Romani & Romani & 74.75 & +1.71 \\
    \midrule
    Second Learners & GECCC & 71.82 & \\
    Second Learners & Second Learners & 72.64 & +0.82 \\
    \bottomrule
    \end{tabular}
\end{table}

\begin{table}[t]
    \centering
    \caption{Results of oversampling factors used for fine-tuning on the GECCC test set. The F\textsubscript{0.5}-score is evaluated on the entire GECCC test set, and NF, NWI, SL, and R are evaluations on the individual domains.}
    \label{tab:finetuning-geccc-oversampling}

    \begin{tabular}{ c|c|cccc }
    \toprule
    Factor & GECCC & NF & NWI & SL & R \\
    \midrule
    0.00 & 71.82 & 72.29 & 75.78 & 68.96 & 71.67 \\ 
    \textbf{0.25} & \textbf{73.52} & \textbf{73.36} & \textbf{77.19} & 71.13 & 73.36 \\ 
    0.50 & 72.96 & 71.93 & 76.09 & 71.11 & 72.80 \\ 
    0.75 & 73.32 & 71.19 & 75.93 & \textbf{71.78} & \textbf{73.55} \\ 
    1.00 & 73.09 & 70.40 & 75.91 & 71.52 & 73.27 \\ 
    \bottomrule
    \end{tabular}
\end{table}

\subsubsection*{Tokenization on Byte-Level or Subwords}

In most cases, GEC models process input text tokenized into subwords. The subword approach combines the benefits of both the word-level representation (short input to the neural network, units of meaning are represented using single vectors) and the character-level representation (even low-frequent and unknown words can be represented). However, in the context of spelling errors, the subword approach faces a new challenge: the representation of a single word can change drastically even with just one character difference. To illustrate, the word ``hypothetically'' is tokenized by the mT5 tokenizer into three subwords ``hypothe'', ``t'', ``ically''. However, when accidentally (and incorrectly) doubling ``p'', the word ``hyppothetically'' tokenizes into four subwords ``hypp'', ``o'', ``thetic'', ``ally''. Therefore, in the presence of input errors, a~model must learn to recognize nearly similar words from vastly different input representations, and it also needs to generate corrections by producing a subword sequence potentially quite different from an input one.

Consequently, we hypothesize that, in the context of grammar error correction, representing the input text using characters or even UTF-8 bytes has the potential to simplify both text encoding and decoding. The performance of the mT5 and ByT5 models presented in Table~\ref{tab:model-comparison-finetuning} and Fig.~\ref{fig:bubble_chart} supports our conjecture, with the ByT5 models of corresponding size significantly surpassing mT5 models.

\begin{table}[t]
    \centering
    \caption{Comparison of different models evaluated on the AKCES and the GECCC datasets (F\textsubscript{0.5}-score).}
    \label{tab:model-comparison-finetuning}

    \def\0{\hphantom{0}}
    \begin{tabular}{l|r|c|c}
    \toprule
    \multicolumn{1}{c|}{Model} & Params & AKCES & GECCC \\
    \midrule
    Rothe~et~al.~(2021)~\textit{\cite{rothe-etal-2021-simple} base} & \textit{580M} & \textit{71.88} & \textit{---} \\
    Rothe~et~al.~(2021)~\textit{\cite{rothe-etal-2021-simple} xxl} & \textit{\kern-.3em 12\,921M} & \textit{83.15}& \textit{---} \\
    Katsumata~\&~Komachi~(2020)~\textit{\cite{katsumata-komachi-2020-stronger}} & \textit{610M} & \textit{73.52}& \textit{---} \\
    
    Náplava~\&~Straka~(2019)~\textit{\cite{naplava-straka-2019-low}} & \textit{210M} & \textit{80.17}& --- \\
    Náplava~et~al.~(2022)~\textit{\cite{naplava-etal-2022-geccc}} & \textit{210M} & \textit{---}& \textit{72.96} \\
    \midrule
    Transformer-small & 10M     & 76.41          & 68.03 \\
    Transformer-base  & 65M     & 81.25          & 73.73 \\
    Transformer-large & 210M    & 81.56          & 73.92 \\
    \midrule
    ByT5-small        & 300M    & 79.89          & 72.56 \\
    ByT5-base         & 582M    & 82.59          & 75.15 \\
    ByT5-large        & 1\,275M & \textbf{84.40} & \textbf{77.01} \\
    \midrule
    mT5-base          & 582M    & 80.08          & 73.46 \\
    mT5-large         & 1\,275M & 82.94          & 76.48 \\
    \bottomrule
    \end{tabular}
\end{table}

\subsubsection*{Model Size}

\begin{figure}[t]
\centering
\includegraphics[width=.8\hsize]{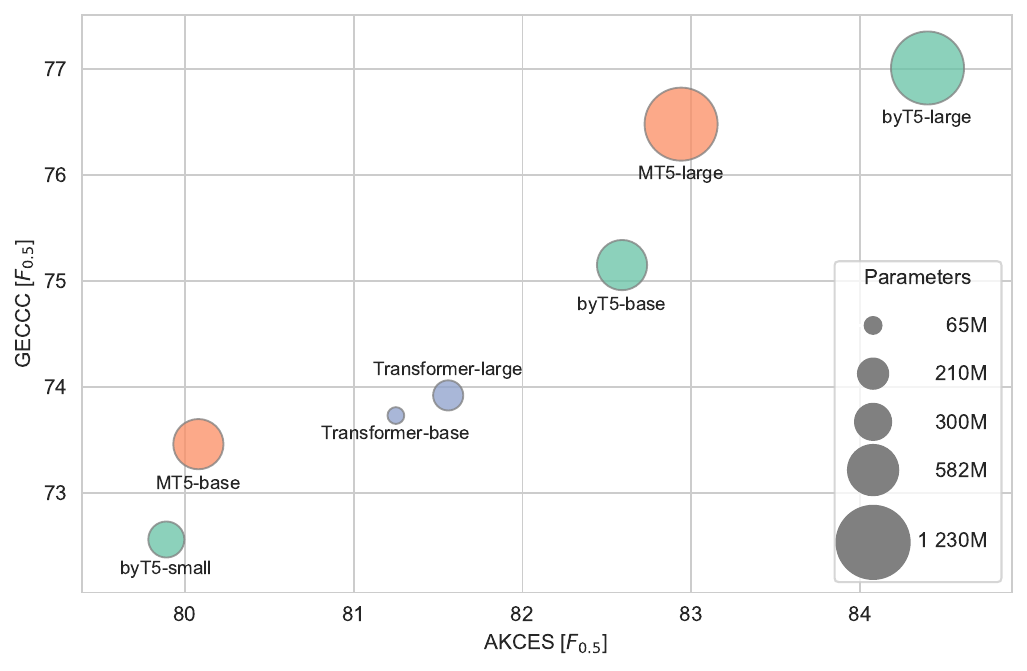}
\caption{Relationship between model size and test F\textsubscript{0.5}-score on both AKCES-GEC (x-axis) and GECCC (y-axis).}
\label{fig:bubble_chart}
\end{figure}

Fig.~\ref{fig:bubble_chart} shows the relationship between model size and test F\textsubscript{0.5}-score on both AKCES-GEC (x axis) and GECCC (y axis). As expected, increasing capacity of a given model improves its results considerably.

\begin{figure}[t!]
\centering
\includegraphics[width=.85\hsize]{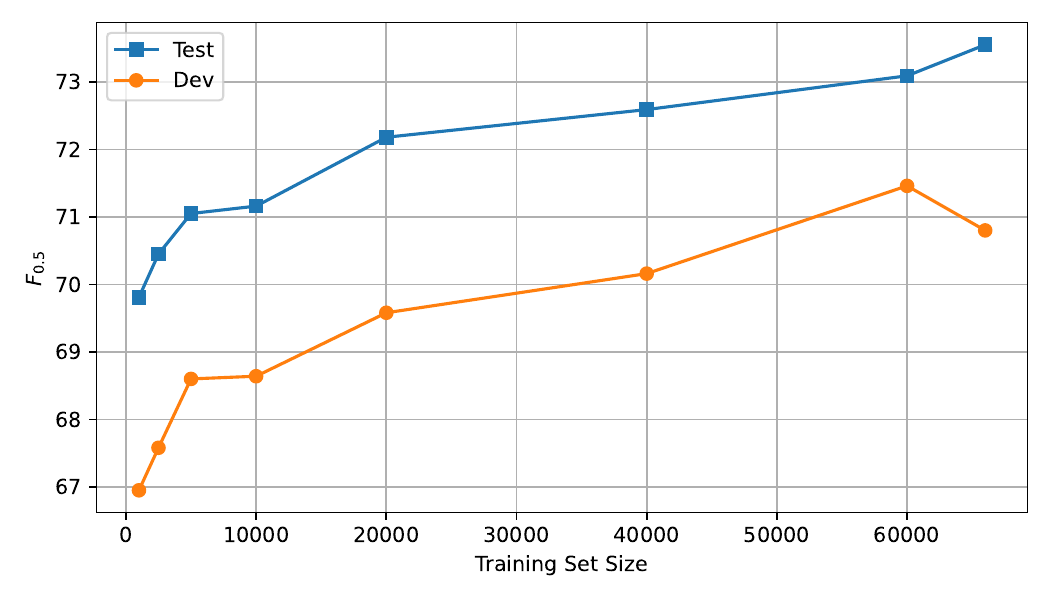}
\caption{The relationship between training set size and the F\textsubscript{0.5}-score during fine-tuning on both the development and test datasets of the GECCC corpus.}
\label{fig:finetuning_performance_vs_training_size}
\end{figure}

\subsubsection*{Data Scaling During Fine-Tuning}

We illustrate the relationship between training set size and the F\textsubscript{0.5}-score during fine-tuning on both the development and test datasets of the GECCC corpus in Fig.~\ref{fig:finetuning_performance_vs_training_size}. Besides the expected performance gain from the increasing data size, we observe only a mild slowdown in performance, indicating that more annotated data could still lead to additional substantial increase in performance.
\section{Fine-tuning LLMs for Czech GEC}

With the availability of large language models pre-trained on vast amount of unsupervised text, it might be possible that fine-tuning such a model solely on manually annotated GEC data could achieve comparable or even better performance compared to pre-training a Transformer from scratch using synthetic data, while being conceptually simpler in the sense of not requiring the synthetic data generation pipeline.

We conduct several experiments with fine-tuned large GPT-like models: Mistral 7B Instruct, and Llama 3.1 8B. Both these LLMs have been pretrained on large data with only a small fraction of Czech.

In the first setting, we use these models as they are, and fine-tune them on the GECCC data. 

In our second setting, we use two proprietary models of Seznam.cz that continued pre-training the LLMs for Czech on large portion of clean Czech data in order to counterweight the low representation of the target language in the original LLMs' pre-training data.\footnote{Similarly to \url{https://laion.ai/blog/leo-lm/} who also continued pre-training LLMs for German.} Then, we also fine-tune these further pre-trained LLMs, resulting in models called Mistral 7B + continued Czech pre-training and Llama 3.1 8B + continued Czech pre-training.\footnote{All the pre-trained LLMs were fine-tuned on GECCC for 10 epochs, each on 4 H100 GPUs with batch size 32 and learning rate 3e-6, cosine scheduler, and a warmup ratio of 0.01. The best checkpoint was selected based on the performance on the GECCC dev set.}

Table~\ref{tab:llm-fine-tuned} summarizes the fine-tuned LLM experiments. We observe that recent Llama 3.1 8B reaches solid results even without synthetic data, and although it was pre-trained with little Czech data; and that continued pre-training focused on the target language (Czech) boosts the LLM for further fine-tuning for the downstream task (GEC).

\begin{table}[t]
    \centering
    \caption{LLMs fine-tuned on the GECCC data either in their original setting or with continued Czech pre-training.}
    \label{tab:llm-fine-tuned}
    \begin{tabular}{ll}
    \toprule
    Model               & F\textsubscript{0.5}-score\\
    \midrule
    Mistral 7B Instruct  & 65.66   \\
    Mistral 7B + continued Czech pre-training   & 72.31\\
    Llama 3.1 8B & 71.49   \\
    Llama 3.1 8B + continued Czech pre-training   & 74.50   \\       
    \bottomrule
\end{tabular}
\end{table}

\section{Evaluation of Czech GEC Approaches}

\begin{table}[t!]
    \centering
    \caption{Czech GEC with LLMs and the Czech GEC tools on 819 (10.36\%) randomly selected test sentences of the GECCC corpus. $\dagger$ Please refer to the disclaimer on possible GECCC train/test data exposure during LLM pretraining.}
    \label{tab:gec_tool_comparison}

    \setlength{\tabcolsep}{3.25pt}
    \begin{tabular}{l|ccccc}
        \toprule
        & \multicolumn{5}{c}{F\textsubscript{0.5}-score} \\
        System & NF & NWI & R & SL & $\sum$ \\
\midrule
\textit{Existing off-the-shelf tools}\\
~~Opravidlo \cite{hlavackova-etal-2022-opravidlo} & 32.95 & 45.97 & 31.51 & 22.13 & 32.76 \\
~~Korektor \cite{richter-etal-2012-korektor} & 36.90 & 24.66 & 48.86 & 54.66 & 44.71 \\
~~Google Docs & 39.56 & 29.03 & 52.23 & 47.13 & 45.45 \\
~~MSWord & 52.25 & 46.20 & 51.63 & 55.22 & 51.54 \\
\midrule
\textit{Non-fine-tuned LLMs}\\
~~DeepSeek R1 70B \cite{deepseekai2025short} zero-shot prompting\textsuperscript{$\dagger$} & 36.06 & 52.34 & 58.46 & 58.11 & 53.58\textsuperscript{$\dagger$} \\
~~GPT4o zero-shot prompting\textsuperscript{$\dagger$} & 59.06 & 78.88 & 77.16 & \textbf{75.64} & 74.60\textsuperscript{$\dagger$} \\
\midrule
\textit{Fine-tuned LLMs}\\
~~Mistral 7B Instruct & 53.30 & 67.03 & 69.71 &  63.67 & 65.39 \\
~~Mistral 7B + cont. Czech pre-training & 62.50 & 77.52  & 75.42 & 72.97 & 73.81 \\
~~Llama 3.1 8B  & 63.73 & 75.41 & 76.41 & 71.69 & 73.54 \\
~~Llama 3.1 8B + cont. Czech pre-training & \textbf{73.39} & 77.24 & 77.00 & 74.99  & 76.18 \\
\midrule
\textit{Specialized GEC systems} \\
~~Náplava et al. (2022) \cite{naplava-etal-2022-geccc} Synthetic & 45.92 & 38.14 & 51.14 & 61.79 & 51.81 \\
~~Náplava et al. (2022) \cite{naplava-etal-2022-geccc} AG-finetuned & 66.45 & 55.02 & 74.39 & 71.81 & 69.82 \\
~~Náplava et al. (2022) \cite{naplava-etal-2022-geccc} GECCC-finetuned & 73.15 & 70.95 & 77.17 & 74.64 & 74.68 \\
\midrule
~~\textbf{Ours (ByT5-large)} & 70.82 & \textbf{82.15} & \textbf{77.45} & 75.38 & \textbf{77.34} \\
    \bottomrule
    \end{tabular}
\end{table}

A comparison of our system with the available off-the-shelf Czech spelling and grammar checking tools (Sec.~\ref{sec:related-work}), as well as selected LLMs, both with and without further fine-tuning, is given in Table~\ref{tab:gec_tool_comparison}. As some of the grammar checkers do not offer automatic batch processing or a REST API, the text had to be processed manually. Therefore, we selected random 10\% of the GECCC test sentences (819 sentences), with representative sampling from each of the four GECCC domains.

Data exposure disclaimer: Since the Czech GEC training, development, and even test data have been freely available online since 2019 (AKCES-GEC) and 2022 (GECCC), and the training corpora of large language models (LLMs) are typically undisclosed, it is impossible to determine whether the evaluation setting is genuinely zero-shot, that is, to what extent the GECCC data may have been seen during pretraining. More concerningly, the test data itself may have been included in the LLMs’ training sets.

As Table~\ref{tab:gec_tool_comparison} shows, the available common grammar checking tools struggle with the informal web discussion texts domain written by native speakers (NWI, Natives Web Informal).
Another interesting point is that we encountered issues with interpretation of the zero-shot prompting tools: while DeepSeek appears to perform on par with the best off-the-shelf tools, GPT4o nearly matches our specialized GEC system results. We attribute these interpretation difficulties to potential data exposure and refer to the abovementioned disclaimer.
Fine-tuned LLMs demonstrate performance gains compared to their non-fine-tuned counterparts and seem to excel on the most cleanest domain of the test data, which are Czech essays (NF, Natives Formal).
However, specialized GEC systems based on the Transformer model trained traditionally in two stages still outperform all the other approaches, and our design improvements presented in Section~\ref{sec:results} significantly improve previous results, while being dramatically more efficient compared to the LLMs.

\section{Conclusions}

We presented a Transformer-based grammar error correction system achieving state-of-the-art results in Czech. We supported our design choices with a wide range of experiments, and we present the following insights based on the results:

\begin{itemize}
\item The SYN-v4 corpus proves superior as a clean corpus base for introducing artificial errors for synthetic data generation.
\item A combination of artificial error generation methods, \textbf{M}orphoDiTa, \textbf{A}spell, and \textbf{T}ypical \textbf{E}rrors, dubbed \textbf{MATE}, outperforms each individual one.
\item Careful domain balancing using (over)sampling is beneficial.
\item Models processing input text as a sequence of bytes achieve better results than models processing subwords.
\item Performance improves with data size, and the mild slowdown suggests further annotated data could still yield substantial gains.
\end{itemize}
Our best-performing model surpasses even models that are several orders of magnitude larger, including recently popular large language models. It also outperforms currently available off-the-shelf Czech GEC tools by a wide margin.

\subsubsection*{\ackname}
Our research has been supported by the OP~JAK project CZ.02.01.01/00/23\_020/0008518 of the Ministry of Education, Youth and Sports of the Czech Republic and uses data provided by the \hbox{LINDAT/CLARIAH-CZ} \hbox{Research} Infrastructure (https://lindat.cz), supported by the Ministry of Education, Youth and Sports of the Czech Republic (Project No. LM2023062).

\bibliographystyle{splncs04}
\bibliography{tsd25_gec}

\end{document}